\documentclass[10pt,twocolumn]{article}

\usepackage[a4paper, margin=1in]{geometry}

\usepackage[utf8]{inputenc}
\usepackage[T1]{fontenc}
\usepackage{lmodern}
\usepackage{microtype}
\usepackage{graphicx}
\usepackage{booktabs}
\usepackage{amsmath,amssymb,amsfonts}
\usepackage{mathtools}
\usepackage{hyperref}
\usepackage{url}
\usepackage{xcolor}
\usepackage{threeparttable}
\usepackage{multirow}
\usepackage{makecell}
\usepackage{stfloats}
\usepackage{algorithm}
\usepackage[noend]{algpseudocode}
\usepackage{natbib}
\usepackage{enumitem}
\usepackage{adjustbox}           

\algnewcommand{\LineComment}[1]{\State \(\triangleright\) \textit{#1}}

\title{The Geometry of Compromise: Unlocking Generative Capabilities\\via Controllable Modality Alignment}

\author{
Hongyuan Liu\textsuperscript{1} \quad
Qinli Yang\textsuperscript{1} \quad
Wen Li\textsuperscript{2} \quad
Zhong Zhang\textsuperscript{1} \quad
Jiaming Liu\textsuperscript{1} \\
Wei Han\textsuperscript{1} \quad
Zhili Qin\textsuperscript{1} \quad
Jinxia Guo\textsuperscript{1} \quad
Junming Shao\textsuperscript{1} \\[6pt]
\textsuperscript{1}University of Electronic Science and Technology of China \quad
\textsuperscript{2}University of Bristol
}

\date{}

\begin{document}

\maketitle

\begin{abstract}
Vision-Language Models (VLMs) such as CLIP learn a shared embedding space for images and text, yet their representations remain geometrically separated, a phenomenon known as the modality gap. This gap limits tasks requiring cross-modal interchangeability, such as captioning and joint clustering. Existing post-processing approaches can partially improve cross-modal compatibility; however, we show through geometric analysis that they primarily reduce the global centroid offset while leaving the underlying distributional mismatch intact. We decompose the modality gap into a Centroid Gap and a Distribution Gap, and demonstrate that the Distribution Gap is the true predictor of cross-modal task quality ($R^2 = 0.986$), whereas the commonly used Raw Gap is misleading ($R^2 = 0.691$).
Motivated by this observation, we propose TPC-CMA (Three-Phase Curriculum for Cross-Modal Alignment), a fine-tuning framework that explicitly reduces both components. The proposed CMA jointly mitigates centroid offsets and reshapes the distributional structure, while a three-phase curriculum with gradient-aware scheduling progressively introduces alignment during training to enable stable optimization.
Experiments demonstrate that our method significantly improves cross-modal alignment. With $\alpha_{\text{target}}{=}0.05$, the modality gap is reduced by 66.6\% with only 4.84\% accuracy drop. Under stronger alignment ($\alpha_{\text{target}}{=}0.5$), the gap is reduced by 82.3\%, clustering ARI improves from 0.318 to 0.516, and captioning CIDEr increases by 57.1\% over the original model. Our code and pre-trained models will be made publicly available upon acceptance.
\end{abstract}


\section{Introduction}

\begin{figure}[t]
  \centering
  \includegraphics[width=0.95\linewidth]{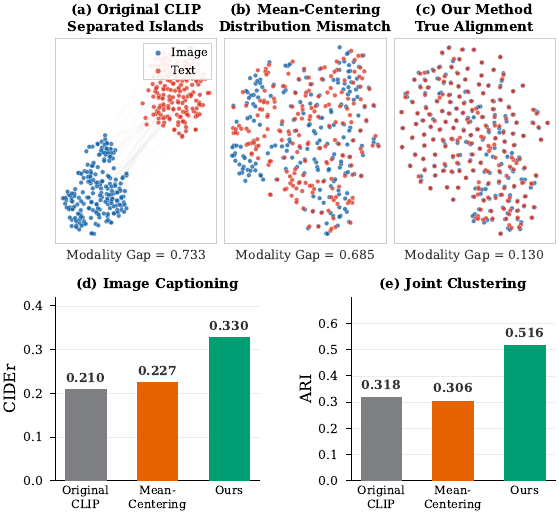}
  \caption{The t-SNE visualization of multimodal features (top) and downstream performance of different methods (bottom). (a)~Original CLIP: image and text embeddings form separated islands. (b)~Mean-Centering\cite{liang2022mind, grclip2025, i0t2025, yamashita2025bridging}: centroids overlap but distributional mismatch persists. (c)~Our alignment method (TPC-CMA, $\alpha_{\text{target}}{=}0.5$): true semantic interleaving. Each panel reports the Modality Gap, defined as the residual cosine distance after centroid alignment (see Section~\ref{sec:gap_decomp}); lower values indicate better structural alignment. \textbf{Bottom}: downstream impact on two zero-shot tasks. (d)~Image captioning CIDEr improves from 0.210 to 0.330. (e)~Joint clustering ARI improves from 0.318 to 0.516.}
  \label{fig:teaser}
\end{figure}

Vision-Language Models (VLMs) such as CLIP~\cite{radford2021learning}, ALIGN~\cite{jia2021align}, SigLIP~\cite{zhai2023sigmoid}, and their successors~\cite{li2023blip2,yu2022coca,sun2024evaclip,cherti2023openclip,chen2025internvl,mu2022slip} have become foundational to modern multimodal AI, enabling powerful zero-shot transfer~\cite{zhou2022coop,gao2024clipadapter,zhai2022lit} via a joint image-text embedding space.

Despite their success, a persistent modality gap exists: image and text embeddings occupy disjoint regions of the hypersphere~\cite{liang2022mind,fahim2024contrastive,schrodi2024twoeffects,ramasinghe2024accept,huang2025mind}. While this gap has limited impact on zero-shot classification, it restricts tasks that require cross-modal feature interchangeability. As shown in Figure~\ref{fig:teaser}(d) and (e), directly feeding image embeddings into a text decoder results in poor captioning performance, and joint image-text clustering yields limited semantic grouping quality. These results indicate that the two modalities fail to form a unified representation space, limiting the applicability of VLMs in generation~\cite{grassucci2026closing,li2023decap,mokady2021clipcap,tewel2022zerocap,zeng2024meacap}, reasoning~\cite{xu2024demystifying}, and broader multimodal understanding tasks~\cite{liu2023llava}.

To mitigate this issue, methods that operate on pretrained embeddings have been widely explored due to their practical advantages. These approaches avoid expensive retraining and preserve the generalization ability of existing VLMs, and can partially improve cross-modal compatibility, as reflected by measurable gains in downstream tasks (Figure~\ref{fig:teaser}(d)(e)). However, the improvements remain limited, especially for tasks requiring deeper cross-modal interaction. This observation suggests that current approaches address only a limited aspect of the underlying modality gap.

In this work, we analyze this limitation from a geometric perspective. We use representative approaches, such as Mean-Centering~\cite{liang2022mind, grclip2025, i0t2025, yamashita2025bridging}, as a diagnostic tool to examine how cross-modal alignment is achieved in practice. As shown in Figure~\ref{fig:teaser}(a), image and text embeddings form two separated regions in the feature space, indicating a fundamental structural discrepancy. We decompose the modality gap into two components: a Centroid Gap, which captures the global shift between modalities, and a Distribution Gap, which reflects differences in their underlying geometric structure. Through this lens, we find that existing approaches primarily reduce the Centroid Gap while leaving the Distribution Gap largely unchanged. This leads to \emph{superficial} alignment as illustrated in Figure~\ref{fig:teaser}(b), where embeddings become closer in position but remain structurally mismatched.
We argue that effective alignment requires consistent geometric structure across modalities.

These observations indicate that resolving the modality gap requires modifying the feature geometry itself, rather than only applying global transformations. To address this limitation, we propose the Three-Phase Curriculum for Cross-Modal Alignment (\textbf{TPC-CMA}), a framework that explicitly reduces both centroid and distribution gaps progressively, which enables an effective geometric-consistent cross-modal alignment as shown in Figure~\ref{fig:teaser}(c). We introduce a Cross-Modal Alignment (CMA) loss that replaces cross-modal repulsion with intra-modal geometry matching, encouraging consistent structures across modalities.
We further design a three-phase curriculum strategy with gradient-aware scheduling~\cite{chen2018gradnorm} that gradually transitions from the original CLIP objective to the alignment objective, enabling stable adaptation of pretrained representations. Extensive experiments demonstrate that our method significantly improves tasks requiring structural cross-modal alignment, while maintaining strong zero-shot performance.

The summary of our contributions is as follows:
\begin{enumerate}
    \item We propose TPC-CMA, a cross-modal alignment framework that reduces the modality gap in a progressive manner.
    \item We introduce a CMA loss that explicitly addresses both centroid and distribution discrepancies by replacing cross-modal repulsion with intra-modal geometry matching, enabling structurally consistent cross-modal alignment.
    \item We design a TPC strategy with gradient-aware ramp-up~\cite{chen2018gradnorm} that gradually transitions from the original CLIP objective to the alignment objective, enabling stable adaptation of pretrained feature geometry without abrupt disruption.
    \item We demonstrate that Distribution Gap, not the commonly used Raw Gap, is a near-perfect predictor of cross-modal task quality ($R^2 = 0.986$ vs.\ $0.691$). Extensive experiments show up to 82.3\% gap reduction, substantial gains in clustering (ARI: 0.318 $\to$ 0.551) and captioning (CIDEr +57.1\%), while maintaining strong zero-shot performance.
\end{enumerate}
\section{Preliminary}
\label{sec:preliminary}
\label{sec:analysis}
\subsection{Pattern Analysis of CLIP}
\label{sec:root_cause}

CLIP~\cite{radford2021learning} learns a shared embedding space via contrastive learning. Given an image encoder $f_I$ and a text encoder $f_T$, inputs are mapped to $\ell_2$-normalized features $\mathbf{v}_i = f_I(x_i)$ and $\mathbf{t}_i = f_T(y_i)$, with similarity defined as $s(\mathbf{v}_i, \mathbf{t}_j) = \mathbf{v}_i^\top \mathbf{t}_j$. The model is trained using a symmetric InfoNCE loss~\cite{oord2018infonce}:
\begin{equation}
\mathcal{L}_{\text{CLIP}} = -\frac{1}{2N}\sum_{i=1}^{N}\left[\log\frac{e^{\tau \mathbf{v}_i^\top \mathbf{t}_i}}{\sum_{j}e^{\tau \mathbf{v}_i^\top \mathbf{t}_j}} + \log\frac{e^{\tau \mathbf{t}_i^\top \mathbf{v}_i}}{\sum_{j}e^{\tau \mathbf{t}_i^\top \mathbf{v}_j}}\right]
\label{eq:clip}
\end{equation}
where $\tau$ is a learnable temperature. This objective can be decomposed into an attractive term $\mathcal{L}_{\text{align}}$ and a repulsive term $\mathcal{L}_{\text{oppose}}$. Following Shi et al.~\cite{shi2023towards}, the image-to-text direction decomposes into:
\begin{align}
\mathcal{L}_{i2t} &= \underbrace{-\frac{1}{N}\sum_{i=1}^{N} \tau\, \mathbf{v}_i^\top \mathbf{t}_i}_{\mathcal{L}_{\text{align}}:\;\text{attraction}} \nonumber \\
&\;+\; \underbrace{\frac{1}{N}\sum_{i=1}^{N} \log\sum_{j=1}^{N} e^{\tau\, \mathbf{v}_i^\top \mathbf{t}_j}}_{\mathcal{L}_{\text{oppose}}:\;\text{repulsion}}
\label{eq:decompose}
\end{align}
with an analogous decomposition for the text-to-image direction. Since each positive pair is contrasted against $N{-}1$ negatives, the cumulative repulsive gradient dominates, introducing a structural asymmetry between attraction and repulsion. As a result, CLIP representations exhibit a characteristic pattern: matched pairs are locally aligned, while image and text embeddings remain globally separated into distinct regions on the hypersphere.

From this perspective, the modality gap can be attributed to the intrinsic geometry induced by the contrastive objective rather than initialization~\cite{liang2022mind}. We provide a theoretical analysis in Appendix~A showing that the opposition term encourages the two modalities to separate in the feature space, leading to an antipodal tendency. This asymmetry explains the coexistence of local alignment and global separation, which we further analyze in the following section.

\subsection{Modality Gap Decomposition}
\label{sec:gap_decomp}
To better understand this phenomenon, we analyze the modality gap from two complementary geometric perspectives.
Given $N$ image-text pairs with $\ell_2$-normalized image embeddings $\mathbf{v}_i$ and text embeddings $\mathbf{t}_i$, the standard modality gap is defined as $\text{Gap} = 1 - \frac{1}{N}\sum_{i} \mathbf{v}_i^\top \mathbf{t}_i$, which we refer to as the Raw Gap, conflates two geometrically distinct phenomena~\cite{role2025fill}. We decompose it into (formal definitions in Appendix~B):

\begin{itemize}
\item \textbf{Centroid Gap} ($\mathcal{G}_C$): $\|\bar{\mathbf{v}} - \bar{\mathbf{t}}\|_2$, where $\bar{\mathbf{v}} = \frac{1}{N}\sum_i \mathbf{v}_i$ and $\bar{\mathbf{t}} = \frac{1}{N}\sum_i \mathbf{t}_i$ are the modal centroids.
\item \textbf{Distribution Gap} ($\mathcal{G}_D$): the residual gap after centroid alignment, measured as the average cosine distance between centered embeddings $\hat{\mathbf{v}}_i = (\mathbf{v}_i - \bar{\mathbf{v}})/\|\mathbf{v}_i - \bar{\mathbf{v}}\|$ and $\hat{\mathbf{t}}_i = (\mathbf{t}_i - \bar{\mathbf{t}})/\|\mathbf{t}_i - \bar{\mathbf{t}}\|$.
\end{itemize}

$\mathcal{G}_C$ and $\mathcal{G}_D$ capture complementary geometric aspects of the modality gap rather than forming a strict additive decomposition. Intuitively, $\mathcal{G}_C$ measures how far apart the two modal clouds are, while $\mathcal{G}_D$ characterizes how differently they are structured. Effective cross-modal alignment requires reducing both components, as correcting centroid shift alone does not resolve structural mismatch.
\begin{figure*}[t]
  \centering
  \includegraphics[width=0.92\textwidth]{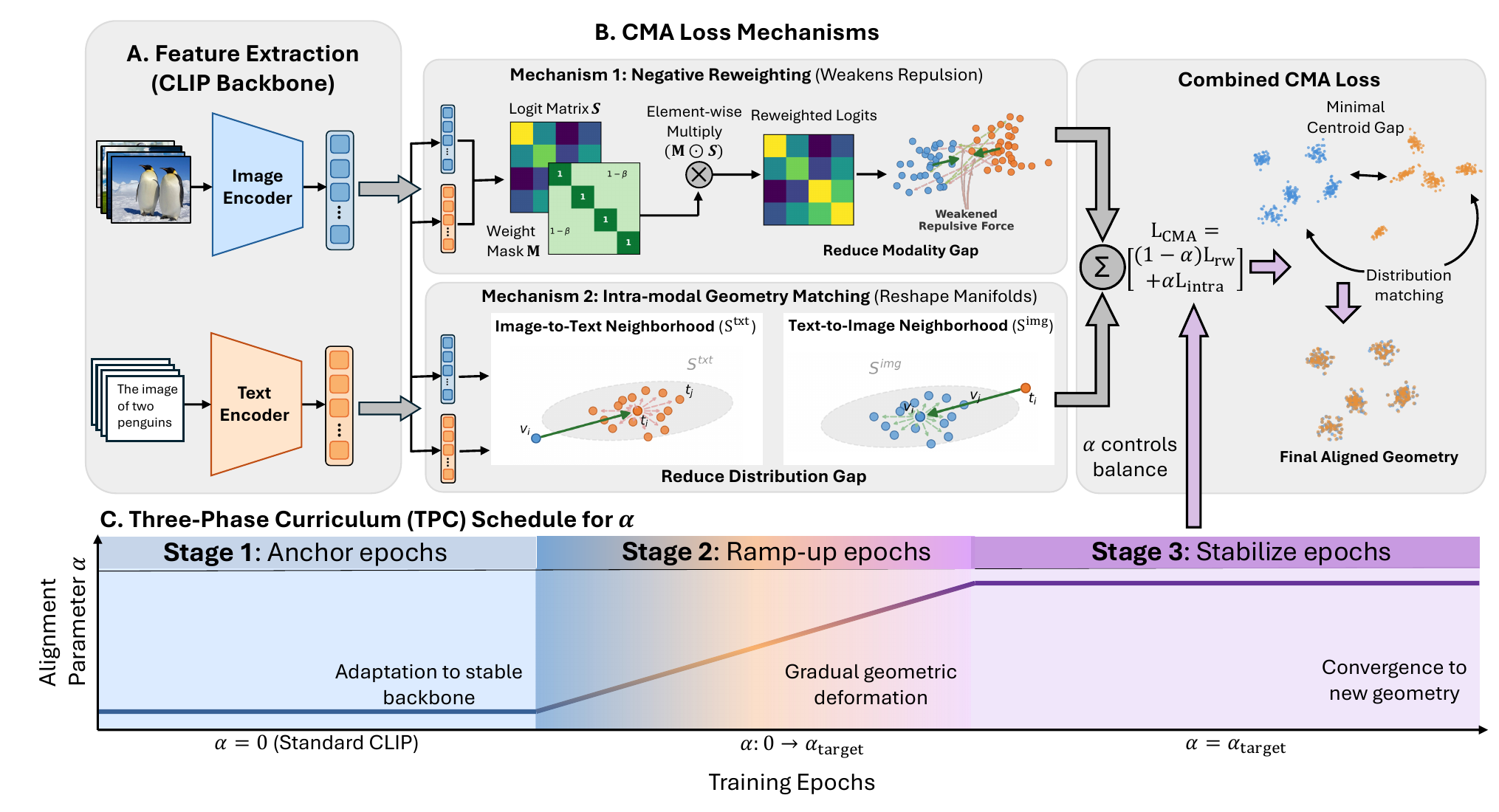}
  \caption{Overview of TPC-CMA. \textbf{A}: CLIP backbone extracts image and text embeddings. \textbf{B}: CMA Loss combines Negative Reweighting (reduces Centroid Gap) and Intra-modal Geometry Matching (reduces Distribution Gap) via $\alpha$. \textbf{C}: Three-Phase Curriculum schedules $\alpha$ from 0 to $\alpha_{\text{target}}$ across Anchor, Gradient-aware Ramp-up, and Stabilize stages, with the transition speed dynamically modulated by the observed gradient dynamics between loss terms.}
  \label{fig:method}
\end{figure*}

\begin{figure}[t]
  \centering
  \includegraphics[width=1.0\linewidth]{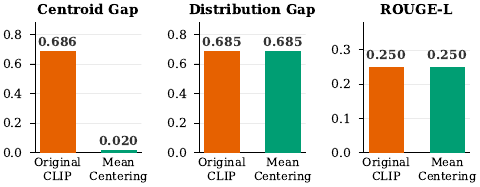}
  \caption{Gap decomposition under Mean-Centering. Despite a 97\% Centroid Gap reduction, both Distribution Gap and ROUGE-L remain unchanged, showing that correcting the centroid offset alone does not improve cross-modal compatibility.}
  \label{fig:gap_decomp_bar}
\end{figure}

\textbf{Analysis Results}: Figure~\ref{fig:gap_decomp_bar} applies this decomposition to Mean-Centering. While the Centroid Gap drops by 97\% (0.686 $\to$ 0.020), the Distribution Gap remains unchanged (0.685). We provide a theoretical analysis in Appendix~B (Proposition~B.1) supporting that translation-based methods preserve $\mathcal{G}_D$ by shifting centroids without altering relative deviations, and ROUGE-L remains at 0.250, confirming no downstream improvement.
This shows that post-processing methods reduce global offsets but fail to address structural discrepancies. Since cross-modal interchangeability is governed by distributional consistency, the unresolved distribution gap becomes the primary bottleneck, motivating alignment strategies that directly enforce structural consistency in the embedding space.

\section{Methodology}
\label{sec:method}

Bridging the modality gap requires addressing the structural discrepancy between modalities, particularly the distributional mismatch that existing approaches do not resolve. Our approach combines a Cross-Modal Alignment (CMA) mechanism to reduce both centroid and distributional gaps (Section~\ref{sec:cma_loss}), with a three-phase training strategy (Section~\ref{sec:tpc}) that gradually introduces alignment during optimization to ensure stable convergence.

\subsection{Overview}
\label{sec:overview}
As illustrated in Figure~\ref{fig:method}, our framework consists of three parts. In Part A, image-text pairs are processed by a CLIP backbone to obtain feature embeddings. In Part B, we introduce a CMA module that reduces the modality gap by addressing both centroid offsets and distributional discrepancies. In Part C, we design a TPC that progressively introduces the alignment objective during training to ensure stable and effective convergence.

\subsection{Cross-Modal Alignment}
\label{sec:cma_loss}

Section~\ref{sec:root_cause} shows that the modality gap arises from both centroid offset and distributional mismatch. Existing methods typically address only one aspect, leading to limited improvement in cross-modal alignment.
To overcome this limitation, we propose a CMA mechanism that jointly targets both components through two complementary designs: (1) Negative Reweighting, which reduces the Centroid Gap, and (2) Intra-modal Geometry Matching, which reshapes the distributional structure to reduce the Distribution Gap.

\subsubsection{Mechanism 1: Negative Reweighting}
\label{sec:neg_rw}

As shown in Eq.~\ref{eq:decompose}, the opposition term $\mathcal{L}_{\text{oppose}}$ accumulates repulsive gradients from all $N{-}1$ unmatched pairs, pushing the two modal distributions toward antipodal regions and sustaining the Centroid Gap.

Specifically, to mitigate this effect, we propose Negative Reweighting, which weakens the contribution of negative pairs in the cross-modal similarity matrix. Let $\mathbf{S} \in \mathbb{R}^{N \times N}$ be the cross-modal logit matrix with $S_{ij} = \tau \mathbf{v}_i^\top \mathbf{t}_j$. We define a weight mask $\mathbf{M}$:
\begin{equation}
M_{ij} = \begin{cases} 1 & \text{if } i = j \\ 1 - \beta & \text{if } i \neq j \end{cases}
\label{eq:weight}
\end{equation}
where $\beta = 0.05\alpha$ controls the strength of reweighting in Eq.~\ref{eq:total_expanded}. We set the coefficient to 0.05 so that $\beta$ remains small (at most 0.05), ensuring that individual negative logits are only mildly suppressed while the cumulative effect over $N{-}1$ negatives is substantial. The reweighted logits $\mathbf{M} \odot \mathbf{S}$ are used in place of $\mathbf{S}$ when computing the cross-entropy loss:
\begin{equation}
\mathcal{L}_{\text{rw}} = \text{CE}(\mathbf{M} \odot \mathbf{S},\; \mathbf{y}) + \text{CE}(\mathbf{M} \odot \mathbf{S}^\top,\; \mathbf{y})
\label{eq:rw}
\end{equation}
where $\text{CE}$ denotes the cross-entropy loss, $\odot$ denotes element-wise (Hadamard) product, and $\mathbf{y} = (1, 2, \ldots, N)$ are the ground-truth labels.
Geometrically, the reduced denominator weakens the gradient on negative pairs, allowing the centroid gap to shrink while still matching pairs correctly. However, reweighting only removes the repulsive barrier without reshaping the distributions.
\subsubsection{Mechanism 2: Intra-modal Geometry Matching}

As shown in Section~\ref{sec:gap_decomp}, even after the Centroid Gap is reduced, the Distribution Gap remains unchanged because the two modalities retain different manifold shapes. Negative Reweighting removes the repulsive barrier but does not prescribe how the underlying manifold geometry should be restructured to achieve distributional consistency.

To close the Distribution Gap, we need to reshape the manifolds so that paired embeddings occupy corresponding positions within their respective distributions. We therefore propose Intra-modal Geometry Matching (IGM), which replaces the cross-modal comparison (``is this image close to its text among all texts?'') with an intra-modal comparison (``is this image's paired text closer than all other texts to this text?''), forcing the two manifolds to adopt geometrically consistent and matching structures.

Specifically, we construct two auxiliary logit matrices:
\begin{equation}
\hat{S}^{(\text{txt})}_{ij} = \begin{cases} \tau\, \mathbf{t}_i^\top \mathbf{v}_i & \text{if } i = j \\ \tau\, \mathbf{t}_i^\top \mathbf{t}_j & \text{if } i \neq j \end{cases}
\label{eq:soft_txt}
\end{equation}
\begin{equation}
\hat{S}^{(\text{img})}_{ij} = \begin{cases} \tau\, \mathbf{v}_i^\top \mathbf{t}_i & \text{if } i = j \\ \tau\, \mathbf{v}_i^\top \mathbf{v}_j & \text{if } i \neq j \end{cases}
\label{eq:soft_img}
\end{equation}

Consider row $i$ in $\hat{S}^{(\text{txt})}$: off-diagonal entries $\tau\, \mathbf{t}_i^\top \mathbf{t}_j$ capture the intra-modal structure of text $i$, while the diagonal $\tau\, \mathbf{t}_i^\top \mathbf{v}_i$ measures cross-modal similarity to the paired image. Cross-entropy forces the paired image to rank higher than all other texts:
\begin{equation}
\mathcal{L}_{\text{intra}} = \text{CE}(\hat{\mathbf{S}}^{(\text{txt})},\; \mathbf{y}) + \text{CE}(\hat{\mathbf{S}}^{(\text{img})},\; \mathbf{y})
\label{eq:intra}
\end{equation}

By minimizing $\text{CE}(\hat{\mathbf{S}}^{(\text{txt})}, \mathbf{y})$, each image $\mathbf{v}_i$ is forced to be closer to its paired text $\mathbf{t}_i$ than any other text, effectively ``inserting'' each image into the structure of its paired modality. The symmetric term $\text{CE}(\hat{\mathbf{S}}^{(\text{img})}, \mathbf{y})$ does the same in reverse. Over the full batch, this bidirectional matching reshapes both manifolds to adopt the same structure, thereby closing the Distribution Gap. Unlike the standard CLIP loss that compares across modalities, $\mathcal{L}_{\mathrm{intra}}$ compares within a single modality, replacing cross-modal opposition with intra-modal uniformity terms \cite{zbontar2021barlow,bardes2022vicreg,grill2020byol} whose optima are consistent with alignment (Appendix~C, Proposition~C.1).
\subsubsection{Combined CMA Loss}

The two mechanisms are combined with the alignment parameter $\alpha \in [0, 1]$, which controls the trade-off between standard contrastive learning and geometry alignment. Let $s_{ij}^{vt}{=}\tau\mathbf{v}_i^\top\mathbf{t}_j$, $s_{ij}^{tt}{=}\tau\mathbf{t}_i^\top\mathbf{t}_j$, $s_{ij}^{vv}{=}\tau\mathbf{v}_i^\top\mathbf{v}_j$, and $\beta{=}0.05\alpha$. The overall CMA loss is defined as:
{\footnotesize
\begin{alignat}{2}
&\mathcal{L}_{\text{CMA}} = \tfrac{1}{2}\big[(1 {-} \alpha)\,\mathcal{L}_{\text{rw}} + \alpha\,\mathcal{L}_{\text{intra}}\big] \nonumber \\
&= -\frac{1}{2N}\sum_{i}\Bigg[(1{-}\alpha)\!\bigg(\log\frac{e^{s_{ii}^{vt}}}{e^{s_{ii}^{vt}} {+} \smash{\underset{j\neq i}{\textstyle\sum}}e^{(1-\beta)s_{ij}^{vt}}} \nonumber \\
&\quad + \log\frac{e^{s_{ii}^{vt}}}{e^{s_{ii}^{vt}} {+} \smash{\underset{j\neq i}{\textstyle\sum}}e^{(1-\beta)s_{ji}^{vt}}}\bigg) \nonumber \\
&\;+ \alpha\!\bigg(\log\frac{e^{s_{ii}^{vt}}}{e^{s_{ii}^{vt}} {+} \smash{\underset{j\neq i}{\textstyle\sum}}e^{s_{ij}^{tt}}} \nonumber \\
&\quad + \log\frac{e^{s_{ii}^{vt}}}{e^{s_{ii}^{vt}} {+} \smash{\underset{j\neq i}{\textstyle\sum}}e^{s_{ij}^{vv}}}\bigg)\Bigg] \label{eq:total_expanded}
\end{alignat}
}
where $s_{ij}^{vt}{=}\tau\mathbf{v}_i^\top\mathbf{t}_j$ denotes the scaled cross-modal similarity, $s_{ij}^{tt}{=}\tau\mathbf{t}_i^\top\mathbf{t}_j$ and $s_{ij}^{vv}{=}\tau\mathbf{v}_i^\top\mathbf{v}_j$ denote intra-modal similarities, $\alpha \in [0,1]$ is the alignment parameter, and $\beta{=}0.05\alpha$ controls the negative reweighting strength. The first two terms correspond to the reweighted contrastive loss $\mathcal{L}_{\text{rw}}$, where $(1{-}\beta)$ down-scales cross-modal negative logits; the last two terms correspond to the intra-modal geometry matching loss $\mathcal{L}_{\text{intra}}$, where negatives come from the same modality. Note that all four terms share the standard log-softmax form $-\log\bigl(e^{s_+}/{\textstyle\sum_j} e^{s_j}\bigr)$; they differ only in which similarities populate the denominator (see Appendix~E for complete pseudocode).
When $\alpha = 0$, $\beta = 0$ so $\mathcal{L}_{\text{rw}}$ reduces to $\mathcal{L}_{\text{CLIP}}$. As $\alpha$ increases, the loss continuously and smoothly transitions from standard contrastive learning toward full intra-modal geometry matching.

\subsection{Three-Phase Curriculum with Gradient-Aware Ramp}
\label{sec:tpc}
While the CMA loss explicitly enforces cross-modal alignment, directly applying the full objective throughout training may lead to suboptimal convergence, as the pretrained representations are not immediately adapted to the modified objective. To address this, we introduce a TPC with gradient-aware scheduling. The curriculum is governed by a weighting parameter \( \alpha \), which controls the strength of the alignment objective. Specifically, TPC gradually increases \( \alpha \) during training, allowing the feature geometry to adapt progressively and converge to a stable optimum.

\paragraph{Stage 1: Anchor ($c$ epochs, $\alpha = 0$).}
The model is trained with standard CLIP loss only, allowing it to adapt to the fine-tuning data while anchoring discriminative structure. All $\alpha$ configurations share the same peak accuracy at the end of this stage.

\paragraph{Stage 2: Gradient-aware Ramp-up ($r$ epochs, $\alpha$: $0 \to \alpha_{\text{target}}$).}
After $c$ epochs of anchoring, the training shifts into the ramp-up phase. A naive approach would increase $\alpha$ at a constant (linear) rate, but the contrastive loss $\mathcal{L}_{\text{rw}}$ and the alignment loss $\mathcal{L}_{\text{intra}}$ exhibit different convergence dynamics: we observe that $\|\nabla\mathcal{L}_{\text{intra}}\|$ dominates early in training (ratio $\approx 2.3{:}1$), indicating gradient conflict, while the ratio converges toward equilibrium ($\approx 1{:}1$) as training progresses. Inspired by multi-task gradient balancing~\cite{chen2018gradnorm}, we modulate the ramp speed based on this convergence signal. Concretely, we maintain a slow EMA $\bar{\ell}_s$ and a fast EMA $\bar{\ell}_f$ of the contrastive loss $\mathcal{L}_{\text{rw}}$. Their ratio $\rho = \bar{\ell}_f / \bar{\ell}_s$ serves as a convergence indicator: $\rho \approx 1$ signals stability, $\rho < 1$ signals that the loss is still dropping (not ready for more alignment), and $\rho > 1$ signals that alignment is hurting discriminative quality. The per-step $\alpha$ update is:
\begin{align}
\Delta\alpha &= \frac{\alpha_{\text{target}} - \alpha}{T - t}
  \cdot \bigl(0.5 + s(\rho)\bigr), \label{eq:ga_ramp} \\
s(\rho) &= \begin{cases}
  \rho & \text{if } \rho \le 1, \\
  2 - \rho & \text{otherwise},
\end{cases}
\quad \rho = \text{clip}\!\left(\frac{\bar{\ell}_f}{\bar{\ell}_s},\; 0,\; 2\right) \nonumber
\end{align}
where $T{-}t$ is the number of remaining ramp steps and the $\text{clip}$ operation bounds $\rho$ to $[0,2]$, ensuring $s(\rho) \in [0,1]$ and the speed factor $(0.5 + s(\rho)) \in [0.5, 1.5]$. The base rate $(\alpha_{\text{target}} - \alpha)/(T-t)$ provides a catch-up mechanism that guarantees $\alpha$ reaches $\alpha_{\text{target}}$ by step $T$: if previous steps were slow, the remaining budget per step automatically increases. The speed factor modulates around the base rate, accelerating when gradients are balanced ($s \to 1$, $\rho \approx 1$) and decelerating when gradient conflict is detected ($s \to 0$).

\paragraph{Stage 3: Stabilize ($h$ epochs, $\alpha = \alpha_{\text{target}}$).}
Finally, the last few epochs stabilize the finetuned feature space. $\alpha$ is held at the target for convergence to a stable geometric configuration.

The full schedule is defined by four values: $(c, r, h, \alpha_{\text{target}})$. We use $c{=}3$, $r{=}5$, $h{=}2$ for a total of 10 epochs across all experiments. The only parameter that users need to choose is $\alpha_{\text{target}}$, which controls the operating point on the Pareto frontier: $\alpha_{\text{target}} \leq 0.05$ for minimal accuracy loss (retrieval, classification); $\alpha_{\text{target}} \geq 0.1$ for deep alignment (clustering, generative tasks). This makes TPC-CMA a protocol: users select $\alpha_{\text{target}}$ based on their downstream task, and the method delivers the optimal geometry.
\section{Experiments}
\label{sec:exp}

\subsection{Experimental Setup}
\label{sec:exp_setup}

\textbf{Datasets and Metrics.}
We use Conceptual Captions 3M (CC3M)~\cite{sharma2018conceptual}, a widely adopted image-text dataset containing approximately 3.3 million web-crawled pairs, as the training set for all fine-tuning methods.
We evaluate on five complementary tasks spanning discriminative, generative, and structural dimensions:
\textbf{(i)} ImageNet~\cite{deng2009imagenet} zero-shot classification (multi-template, consistent with CLIP's original protocol);
\textbf{(ii)} COCO~\cite{lin2014coco} image-text retrieval (5K test split, R@1);
\textbf{(iii)} multi-dataset zero-shot classification (CIFAR-10/100, Food-101, Caltech-101, Flowers-102);
\textbf{(iv)} zero-shot captioning (COCO, 5K images): following DeCap~\cite{li2023decap}, we train a separate lightweight GPT-2~\cite{radford2019gpt2} decoder per model variant on text embeddings and directly feed image embeddings as the decoder prefix at test time. Since the decoder is trained exclusively on text embeddings and tested on image embeddings, CIDEr improvements reflect the degree to which image embeddings have become interchangeable with text embeddings, not the decoder's capacity. We report CIDEr and ROUGE-L;
\textbf{(v)} joint image-text clustering: following the group-wise evaluation protocol of Grassucci et al.~\cite{grassucci2026closing}, we sample 200 classes from ImageNet validation (50 images per class), pair each image with a text prompt ``a photo of a \{classname\}'', combine all image and text embeddings into a single pool (20{,}000 vectors), and run KMeans with $k{=}200$. Ground-truth labels are semantic class identities (shared across modalities). We report V-Measure~\cite{rosenberg-hirschberg-2007-v} and Adjusted Rand Index (ARI) as clustering quality metrics.

\textbf{Implementation Details.}
We fine-tune the pre-trained ViT-B/32~\cite{dosovitskiy2021vit,vaswani2017attention} on CC3M for 10 epochs with a learning rate of $1{\times}10^{-5}$ and an effective batch size of 4096. The learnable temperature $\tau$ is initialized from the pre-trained checkpoint and continues to be optimized during fine-tuning. The three-phase curriculum uses $c{=}3$, $r{=}5$, $h{=}2$ throughout. For generalization, we additionally evaluate ViT-B/16 and ViT-L/14 (see Appendix~F).

\textbf{Baselines.}
We compare against five representative methods mitigating the modality gap with post-processing, fine-tuning, and pre-training stage:
(1)~\textbf{Mean-Centering}~\cite{liang2022mind}, a post-processing translation that subtracts modal centroids;
(2)~\textbf{AlignCLIP}~\cite{alignclip2025} (ICLR'25), which trains a CLIP model from scratch with an alignment objective;
(3)~\textbf{M$^2$-Mix}~\cite{oh2023geodesic} (NeurIPS'23), which generates hard negatives via geodesic interpolation;
(4)~\textbf{CLIP-Refine}~\cite{cliprefine2025} (CVPR'25), which applies random-reference feature alignment during pre-training;
(5)~\textbf{CS-Aligner}~\cite{yin2025distributional} (ICLR'26), which adds a Cauchy--Schwarz divergence term to InfoNCE for distributional alignment.
All fine-tuning and post-processing baselines use the same training data, number of epochs, and backbone architecture for a fair comparison.
\begin{table*}[t]
    \caption{Main comparison across gap reduction, discriminative, generative, and structural tasks.}
    \label{tab:main_results}
    \centering
    \resizebox{\textwidth}{!}{%
    \begin{tabular}{llccccccccc}
      \toprule
      Method & Type & Gap & Gap$\downarrow$\% & $\mathcal{G}_D$ & ImageNet & I2T R@1 & T2I R@1 & V-Measure & CIDEr & ROUGE-L\\
      \midrule
      Original CLIP \cite{radford2021learning} & Pretrained & 0.733 & {-} & 0.685 & \underline{62.62} & \underline{53.42} & 35.36 & 0.769 & 0.210 & 0.250 \\
      Mean-Centering \cite{liang2022mind} & Post-proc & 0.685 & 6.5 & 0.685 & 60.23 & 43.74 & 34.44 & 0.766 & 0.227 & 0.250 \\
      AlignCLIP \cite{alignclip2025}$^{\dagger}$ & From-scratch & 0.353 & 51.8 & 0.550 & 32.79 & 33.64 & 22.28 & 0.749 & 0.112 & 0.244 \\
      M$^2$-Mix \cite{oh2023geodesic} & Fine-tune & 0.710 & 3.1 & 0.637 & 54.24 & 42.98 & 31.00 & 0.762 & 0.197 & 0.239 \\
      CLIP-Refine \cite{cliprefine2025} & Fine-tune & 0.747 & $-$1.9 & 0.647 & \textbf{62.99} & \textbf{53.90} & \textbf{36.31} & 0.765 & 0.236 & 0.265 \\
      CS-Aligner \cite{yin2025distributional} & Fine-tune & 0.734 & $-$0.1 & 0.617 & 61.37 & 49.15 & 32.89 & 0.770 & 0.174 & 0.249 \\
      \midrule
      TPC-CMA $\alpha_{\text{target}}{=}0.01$ & Fine-tune & 0.558 & 23.8 & 0.610 & 60.86 & 50.96 & \underline{35.96} & 0.795 & 0.227 & 0.273 \\
      TPC-CMA $\alpha_{\text{target}}{=}0.05$ & Fine-tune & 0.245 & 66.6 & 0.554 & 57.78 & 48.56 & 35.68 & 0.810 & 0.254 & 0.269 \\
      TPC-CMA $\alpha_{\text{target}}{=}0.3$ & Fine-tune & 0.155 & 78.9 & 0.474 & 55.63 & 46.44 & 35.08 & 0.826 & 0.293 & 0.276 \\
      TPC-CMA $\alpha_{\text{target}}{=}0.5$ & Fine-tune & \underline{0.130} & \underline{82.3} & \underline{0.441} & 54.86 & 45.48 & 34.49 & \textbf{0.849} & \underline{0.330} & \underline{0.288} \\
      TPC-CMA $\alpha_{\text{target}}{=}0.9$ & Fine-tune & \textbf{0.096} & \textbf{87.0} & \textbf{0.392} & 52.58 & 42.60 & 33.38 & \underline{0.845} & \textbf{0.356} & \textbf{0.296} \\
      \bottomrule
    \end{tabular}}

    \vspace{2pt}
    {\footnotesize $^{\dagger}$\,Evaluated with AlignCLIP released checkpoint. Training from scratch is prohibitively expensive yet generally unable to match the discriminative quality of fine-tuning a pre-trained model. \textbf{Bold} indicates the best result and \underline{underline} the second best. The same convention applies to all subsequent tables.}
\end{table*}

\begin{figure}[h]
  \centering
  \includegraphics[width=\linewidth]{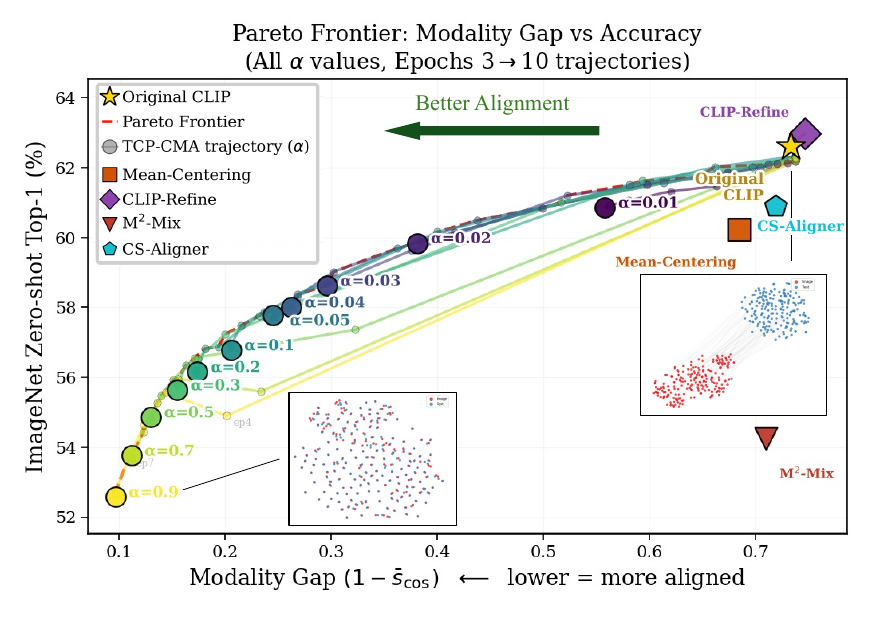}
  \caption{Gap vs.\ Accuracy Pareto Frontier. TPC-CMA forms a smooth efficient frontier that envelopes all selected baselines. AlignCLIP and M$^2$-Mix fall significantly below the frontier, while CLIP-Refine fails to reduce the gap.}
  \label{fig:pareto}
\end{figure}

\subsection{The Pareto Frontier}
\label{sec:main_results}

As shown in Section~\ref{sec:gap_decomp}, existing methods each address only part of the modality gap, leading to limited or one-sided improvements. To demonstrate that TPC-CMA overcomes this limitation by simultaneously reducing both Centroid Gap and Distribution Gap, we compare it against all baselines on five axes: gap reduction, classification, retrieval, clustering, and captioning. Table~\ref{tab:main_results} summarizes the results, and Figure~\ref{fig:pareto} visualizes their gap vs.\ accuracy trade-offs.

Specifically, each baseline occupies a suboptimal position on the Gap vs.\ Accuracy frontier. CLIP-Refine~\cite{cliprefine2025} preserves discriminative quality (ImageNet 62.99\%) but fails to reduce the gap ($-$1.9\%). CS-Aligner~\cite{yin2025distributional} adds a Cauchy--Schwarz divergence penalty that lowers $\mathcal{G}_D$ to 0.617 ($\downarrow$9.9\%) yet leaves Raw Gap unchanged (0.734), exhibiting the same centroid-gap-agnostic pattern as CLIP-Refine. M$^2$-Mix~\cite{oh2023geodesic} reduces the gap by only 3.1\% at a disproportionate accuracy cost (ImageNet $-$8.38\%). AlignCLIP achieves a low gap (0.353) but sacrifices discriminative quality (ImageNet 32.79\%). Mean-Centering reduces the gap by 6.5\% but degrades I2T R@1 by 18.1\%, the superficial alignment identified in Section~\ref{sec:gap_decomp}. We analyze why these methods produce limited improvements through the Distribution Gap lens in Section~\ref{sec:aligned_mode}. In contrast, TPC-CMA's operating points (blue curve in Figure~\ref{fig:pareto}) envelope every baseline, reducing both Raw Gap and $\mathcal{G}_D$ simultaneously. At mild alignment ($\alpha_{\text{target}}{=}0.01$), $\mathcal{G}_D$ drops from 0.685 to 0.610 with less than 2\% ImageNet loss. At strong alignment ($\alpha_{\text{target}}{=}0.5$), $\mathcal{G}_D$ reaches 0.441 ($\downarrow$36\%), CIDEr improves by 57.1\%, and ARI reaches 0.516. Importantly, each $\alpha_{\text{target}}$ simultaneously reports results on all five tasks, confirming that TPC-CMA is a single protocol with a continuously adjustable operating point, not a family of task-specific models.

These results show that the limitations of prior methods are not inherent to the alignment problem itself. TPC-CMA provides a controllable Pareto frontier by jointly addressing both gap components identified in our decomposition analysis (Section~\ref{sec:gap_decomp}).


\begin{table}[h]
  \caption{DeCap zero-shot captioning on COCO. Image embeddings are directly used as the decoder prefix without any learned projection layer.}
  \label{tab:decap}
  \centering
  \resizebox{\columnwidth}{!}{%
  \begin{tabular}{lcccc}
    \toprule
    Method & $\mathcal{G}_D$ & CIDEr & BLEU-4 & ROUGE-L \\
    \midrule
    Original CLIP & 0.685 & 0.210 & 0.029 & 0.250 \\
    Mean-Centering & 0.685 & 0.227 & 0.040 & 0.250 \\
    AlignCLIP & 0.550 & 0.112 & 0.015 & 0.244 \\
    CLIP-Refine & 0.647 & 0.236 & 0.045 & 0.265 \\
    M$^2$-Mix & 0.637 & 0.197 & 0.032 & 0.239 \\
    CS-Aligner & 0.617 & 0.174 & 0.036 & 0.249 \\
    \midrule
    TPC-CMA $\alpha_{\text{target}}{=}0.01$ & 0.610 & 0.227 & 0.050 & 0.273 \\
    TPC-CMA $\alpha_{\text{target}}{=}0.05$ & 0.554 & 0.254 & 0.044 & 0.269 \\
    TPC-CMA $\alpha_{\text{target}}{=}0.3$ & 0.474 & 0.293 & 0.048 & 0.276 \\
    TPC-CMA $\alpha_{\text{target}}{=}0.5$ & 0.441 & \underline{0.330} & \underline{0.057} & \underline{0.288} \\
    TPC-CMA $\alpha_{\text{target}}{=}0.9$ & 0.392 & \textbf{0.356} & \textbf{0.065} & \textbf{0.296} \\
    \bottomrule
  \end{tabular}}
\end{table}

\begin{table}[h]
  \caption{Joint image-text clustering following Grassucci et al.~\cite{grassucci2026closing} (ImageNet val, 200 classes $\times$ 50 imgs, KMeans $k{=}200$).}
  \label{tab:clustering}
  \centering
  \resizebox{\columnwidth}{!}{%
  \begin{tabular}{lccc}
    \toprule
    Method & $\mathcal{G}_D$ & V-Measure & ARI \\
    \midrule
    Original CLIP & 0.685 & 0.769 & 0.318 \\
    Mean-Centering & 0.685 & 0.766 & 0.306 \\
    AlignCLIP & 0.550 & 0.749 & 0.304 \\
    M$^2$-Mix & 0.637 & 0.762 & 0.298 \\
    CLIP-Refine & 0.647 & 0.765 & 0.310 \\
    CS-Aligner & 0.617 & 0.770 & 0.315 \\
    \midrule
    TPC-CMA $\alpha_{\text{target}}{=}0.01$ & 0.610 & 0.795 & 0.330 \\
    TPC-CMA $\alpha_{\text{target}}{=}0.05$ & 0.554 & 0.810 & 0.385 \\
    TPC-CMA $\alpha_{\text{target}}{=}0.3$ & 0.474 & 0.826 & 0.435 \\
    TPC-CMA $\alpha_{\text{target}}{=}0.5$ & 0.441 & \underline{0.849} & \underline{0.516} \\
    TPC-CMA $\alpha_{\text{target}}{=}0.7$ & 0.417 & \textbf{0.863} & \textbf{0.551} \\
    TPC-CMA $\alpha_{\text{target}}{=}0.9$ & 0.392 & 0.845 & 0.492 \\
    \bottomrule
  \end{tabular}}
\end{table}
\subsection{Generative and Structural Tasks}
\label{sec:aligned_mode}

To verify that alignment can substantially improve generative and structural capabilities beyond what the original model achieves, we evaluate TPC-CMA on two tasks that directly depend on cross-modal geometric coherence: zero-shot captioning without projection (Table~\ref{tab:decap}), which tests feature interchangeability, and joint image-text clustering (Table~\ref{tab:clustering}), which tests the quality of unified semantic grouping across modality boundaries.
\begin{figure}[h]
  \centering
  \includegraphics[width=\linewidth]{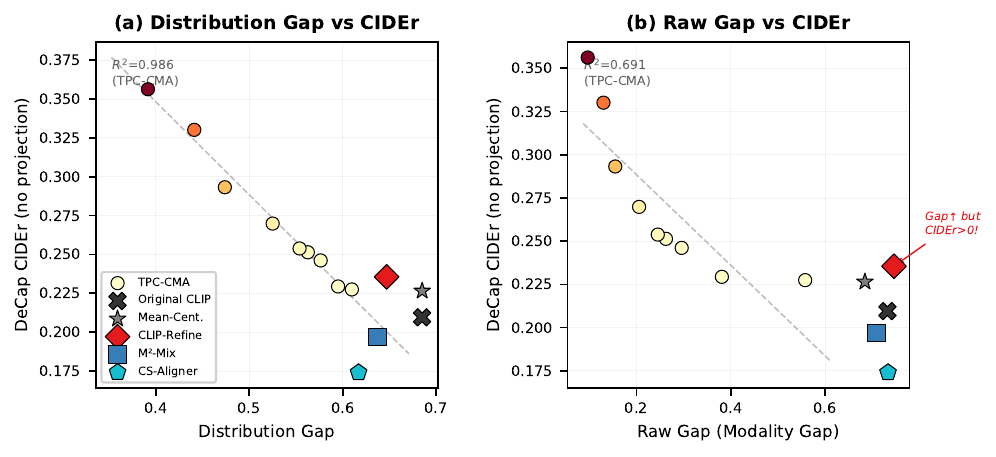}
  \caption{Distribution Gap vs.\ Raw Gap as predictors of DeCap CIDEr score. (a)~Distribution Gap is a near-perfect predictor ($R^2 = 0.986$); (b)~Raw Gap yields a substantially weaker fit ($R^2 = 0.691$).}
  \label{fig:distribgap_cider}
\end{figure}

Specifically, for zero-shot captioning (Table~\ref{tab:decap}), we feed image embeddings directly to a text decoder without DeCap's memory-bank projection, testing cross-modal feature interchangeability. For TPC-CMA, Distribution Gap $\mathcal{G}_D$ decreases monotonically from 0.610 ($\alpha_{\text{target}}{=}0.01$) to 0.392 ($\alpha_{\text{target}}{=}0.9$), and CIDEr increases correspondingly from 0.227 to 0.356 (69.5\% higher than the Original CLIP baseline of 0.210). Figure~\ref{fig:distribgap_cider} formalizes this relationship: $\mathcal{G}_D$ is a near-perfect linear predictor of CIDEr for TPC-CMA ($R^2 = 0.986$), whereas Raw Gap yields a much weaker fit ($R^2 = 0.691$). This confirms that Distribution Gap, not Raw Gap, governs cross-modal interchangeability, validating the decomposition in Section~\ref{sec:gap_decomp}.

The baselines corroborate this finding. Mean-Centering's $\mathcal{G}_D$ remains identical to Original CLIP (0.685), confirming the Section~\ref{sec:gap_decomp} analysis; its marginal CIDEr gain reflects centroid shift alone, not structural improvement. CLIP-Refine's slightly lower $\mathcal{G}_D$ (0.647) correctly predicts its higher CIDEr despite a larger Raw Gap. M$^2$-Mix and AlignCLIP fall below the trend line because degraded representation quality prevents $\mathcal{G}_D$ reduction from translating into proportional CIDEr gains, indicating that $\mathcal{G}_D$ governs interchangeability \emph{given adequate representation quality}. We provide qualitative captioning examples in Appendix~G to illustrate the improvement.

For joint clustering (Table~\ref{tab:clustering}), Original CLIP achieves ARI = 0.318, reflecting limited semantic coherence in joint clustering. TPC-CMA progressively improves clustering quality as $\mathcal{G}_D$ decreases, with ARI rising from 0.330 ($\alpha_{\text{target}}{=}0.01$) to 0.551 ($\alpha_{\text{target}}{=}0.7$), while ARI at $\alpha_{\text{target}}{=}0.9$ shows a decline (0.492) despite further $\mathcal{G}_D$ reduction. We analyze the spectral properties underlying this saturation in Section~\ref{sec:conclusion}. Mean-Centering achieves only ARI = 0.306, barely above Original CLIP despite centroid alignment, confirming that structural reshaping, not just centroid correction, is needed for meaningful improvement in joint clustering quality across modalities.

These task-specific optima underscore the necessity of controllable alignment strength: CIDEr favors high $\alpha_{\text{target}}$, ARI peaks at moderate $\alpha_{\text{target}}$, and discriminative accuracy favors low $\alpha_{\text{target}}$ (visualized in Appendix~H). No single fixed operating point can satisfy all tasks simultaneously, making TPC-CMA's controllable $\alpha_{\text{target}}$ a geometric necessity rather than a mere convenience.

\subsection{Discriminative Tasks}
\label{sec:safe_mode}

\begin{table}[h]
  \caption{COCO retrieval (5K test split).}
  \label{tab:retrieval}
  \centering
  \resizebox{\columnwidth}{!}{%
  \begin{tabular}{lcccc}
    \toprule
    Method & I2T R@1 & I2T R@5 & T2I R@1 & T2I R@5 \\
    \midrule
    Original CLIP & \textbf{53.42} & \textbf{77.18} & 35.36 & 61.52 \\
    Mean-Centering & 43.74 & 70.70 & 34.44 & 60.03 \\
    \midrule
    TPC-CMA $\alpha_{\text{target}}{=}0.01$ & \underline{50.96} & \underline{76.10} & \textbf{35.96} & \textbf{62.16} \\
    TPC-CMA $\alpha_{\text{target}}{=}0.05$ & 48.56 & 74.86 & \underline{35.68} & \underline{61.87} \\
    \bottomrule
  \end{tabular}}
\end{table}

\begin{table}[h]
  \caption{Multi-dataset zero-shot classification.}
  \label{tab:zeroshot}
  \centering
  \resizebox{\columnwidth}{!}{%
  \begin{tabular}{lccccc}
    \toprule
    Method & ImageNet & CIFAR-100 & Food-101 & Caltech & Flowers \\
    \midrule
    Orig.\ CLIP & \textbf{62.62} & \textbf{69.80} & \textbf{79.50} & 85.90 & \textbf{67.20} \\
    Mean-Centering & \underline{60.23} & 67.60 & \underline{78.50} & 83.90 & 62.00 \\
    \midrule
    TPC-CMA $\alpha_{\text{target}}{=}0.01$ & 60.86 & \underline{67.80} & 77.00 & \textbf{87.10} & \underline{64.00} \\
    TPC-CMA $\alpha_{\text{target}}{=}0.05$ & 57.78 & 66.20 & 75.30 & \underline{87.00} & 62.30 \\
    \bottomrule
  \end{tabular}}
\end{table}
To show that mild alignment ($\alpha_{\text{target}} \leq 0.05$) can preserve, or even improve, discriminative quality, we compare TPC-CMA at low $\alpha_{\text{target}}$ against Original CLIP and Mean-Centering on COCO retrieval (Table~\ref{tab:retrieval}) and five zero-shot classification benchmarks (Table~\ref{tab:zeroshot}). We focus on Original CLIP and Mean-Centering as references, since Table~\ref{tab:main_results} shows that the remaining baselines are unsuitable comparisons here: M$^2$-Mix and AlignCLIP suffer severe accuracy degradation (ImageNet $-$8.38\% and 32.79\%, respectively), while CLIP-Refine fails to reduce the gap at all ($-$1.9\% change).

For retrieval (Table~\ref{tab:retrieval}), at $\alpha_{\text{target}}{=}0.01$, T2I R@1 slightly exceeds Original CLIP while I2T R@1 drops by only 2.46 points, far less than Mean-Centering's 9.68-point collapse. For zero-shot classification (Table~\ref{tab:zeroshot}), coarse-grained benchmarks like Caltech-101 \emph{improve} ($85.90 \to 87.10\%$), benefiting from tighter image-to-class-name correspondence, while fine-grained benchmarks like ImageNet show only mild degradation ($-$1.76\% at $\alpha_{\text{target}}{=}0.01$). These results establish that \emph{mild alignment occupies a ``better-than-baseline'' regime for concept-level tasks}, refuting the assumption that gap reduction and discriminative quality are strictly antagonistic.


\subsection{Ablation Studies}
\label{sec:ablation}

\begin{table}
  \caption{Component ablation. w/o Intra applies Negative Reweighting only. CMA-Only applies both CMA mechanisms but with constant $\alpha$ from epoch~1 (no three-phase curriculum).}
  \label{tab:component_ablation}
  \centering
  \resizebox{\columnwidth}{!}{%
  \begin{tabular}{lcccccccc}
    \toprule
    Variant & NR & Intra & TPC & $\alpha$ & \makecell{Modal.\\Gap $\downarrow$} & \makecell{ImageNet\\Top-1} & ARI \\
    \midrule
    w/o NR & & \checkmark & \checkmark & 0.5 & 0.282 & \underline{54.66} & \underline{{0.453}} \\
    w/o Intra & \checkmark & & \checkmark & 0.5 & 0.239 & 53.13 & {0.348} \\
    CMA-Only& \checkmark & \checkmark & & 0.5 & \underline{0.133} & 49.75 & {0.362} \\
    TPC-CMA & \checkmark & \checkmark & \checkmark & 0.5 & \textbf{0.130} & \textbf{54.86} & \textbf{0.516} \\
    \midrule
    w/o NR & & \checkmark & \checkmark & 0.05 & 0.364 & \underline{55.47} & \underline{{0.372}} \\
    w/o Intra & \checkmark & & \checkmark & 0.05 & 0.379 & 55.19 & {0.342} \\
    CMA-Only& \checkmark & \checkmark & & 0.05 & \textbf{0.244} & 51.71 & {0.335} \\
    TPC-CMA & \checkmark & \checkmark & \checkmark & 0.05 & \underline{0.245} & \textbf{57.78} & \textbf{0.385} \\

    \bottomrule
  \end{tabular}}
\end{table}

To isolate the contribution of each component in TPC-CMA, namely Negative Reweighting (NR), Intra-modal Geometry Matching (Intra), and the Three-Phase Curriculum (TPC), we conduct a component removal ablation (Table~\ref{tab:component_ablation}) at two representative $\alpha$ values.

\textbf{Geometry Matching is essential.}
Without Geometry Matching (w/o Intra, $\alpha{=}0.5$), Negative Reweighting alone reduces the gap but ARI reaches only {0.348}. Adding Geometry Matching boosts ARI to 0.516 ($1.5{\times}$), confirming that manifold reshaping, not just repulsion reduction, is essential for structural tasks.

\textbf{Curriculum is essential.}
CMA-Only applies both CMA mechanisms but at a constant $\alpha{=}0.5$ from epoch~1. Despite reaching a nearly identical gap (0.133 vs.\ 0.130), it suffers {$1.4{\times}$} ARI degradation and 5\% ImageNet loss compared to full TPC-CMA. Abrupt alignment destroys the pretrained discriminative structure before the model can adapt, so even a well-closed gap lacks the feature quality for meaningful clustering. The three-phase curriculum prevents this by anchoring features first and increasing alignment only as fast as the observed gradient equilibrium permits.

\textbf{Additional ablations and generalization.}
We further validate curriculum hyperparameters (schedule type and anchor phase length) and confirm that TPC-CMA generalizes across model sizes with consistent gap reduction ratios and no architecture-specific tuning. These results are detailed in Appendix~F.

\section{Discussion}
\label{sec:conclusion}
The preceding experiments reveal task-specific optima at different $\alpha_{\text{target}}$ values, suggesting an intrinsic cost to strong alignment. We examine the spectral properties of the aligned feature space to understand this cost.

\begin{figure}[h]
  \centering
  \includegraphics[width=\linewidth]{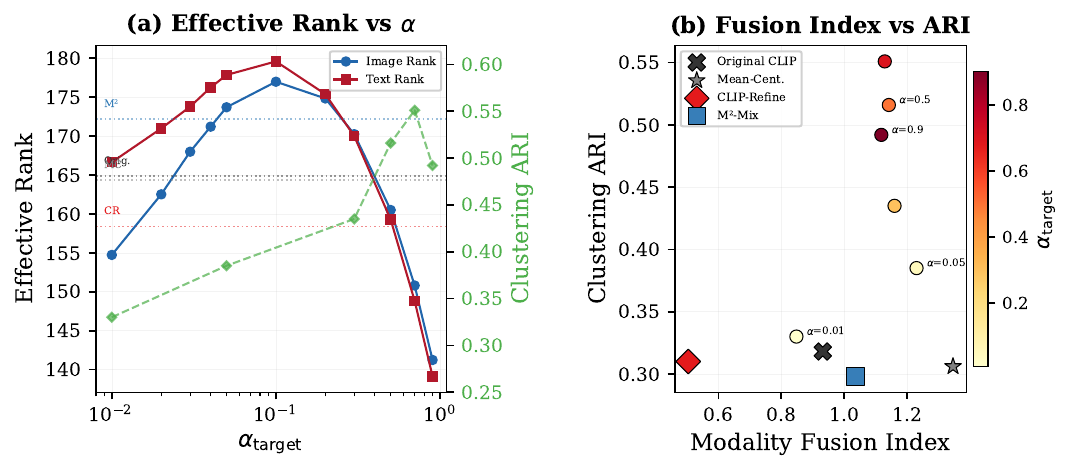}
  \caption{Effective rank dynamics. (a)~Single-modal effective rank vs.\ $\alpha_{\text{target}}$; rank follows an inverted-U, peaking near $\alpha_{\text{target}}{=}0.1$ before declining. Baselines are shown as horizontal dotted lines. (b)~Modality Fusion Index vs.\ ARI.}
  \label{fig:rank_dynamics}
\end{figure}

\textbf{Effective rank and the cost of over-alignment.}
As shown in Figure~\ref{fig:rank_dynamics}(a), the effective rank~\cite{roy2007effrank} of each modality's feature space follows an inverted-U: at moderate alignment, CMA's uniformity component expands feature utilization beyond the original model, but at high $\alpha_{\text{target}}$ alignment pressure compresses features into a lower-dimensional subspace. This rank collapse explains why accuracy drops and ARI saturates under strong alignment.

\textbf{Modality Fusion Index.}
We define FusionIdx = JointRank / mean(ImgRank, TxtRank) to quantify cross-modal complementarity, where a value near 2 indicates orthogonal subspaces and near 1 indicates full overlap. Figure~\ref{fig:rank_dynamics}(b) shows that FusionIdx peaks at $\alpha_{\text{target}}{=}0.05$ and declines as stronger alignment forces both modalities into the same subspace. Notably, CLIP-Refine's FusionIdx of 0.504 (below 1, indicating redundant collapse) explains its failure to improve ARI beyond the Original CLIP baseline (0.310 vs.\ 0.318) despite a low Raw Gap, reinforcing that spectral health, not gap magnitude, determines cross-modal utility.

\textbf{Implications for controllable alignment.}
These spectral dynamics bracket the useful operating range: too low $\alpha_{\text{target}}$ leaves the Distribution Gap intact, while too high triggers rank collapse that erodes representation quality. Since tasks differ in sensitivity to these pressures, TPC-CMA's adjustable $\alpha_{\text{target}}$ is a geometric necessity for effective deployment across diverse tasks.

\section{Related Work}

\subsection{Understanding the Modality Gap.}
Liang et al.~\cite{liang2022mind} first identified the modality gap in CLIP~\cite{radford2021learning}, attributing the separation to a ``cone effect'' from random initialization. Subsequent work challenged this view: Fahim et al.~\cite{fahim2024contrastive} and Shi et al.~\cite{shi2023towards} show the gap is intrinsic to the contrastive objective, while Schrodi et al.~\cite{schrodi2024twoeffects} link it to object bias and information imbalance. Wang and Isola~\cite{wang2020understanding} reveal a tension between alignment and uniformity, extended to multimodal settings by Yin et al.~\cite{yin2026uniformity,yin2025distributional}. Further geometric analyses include similarity structure~\cite{yi2025decipher}, feature norm disparities~\cite{yaras2025explaining}, and global/residual decomposition~\cite{role2025fill} related to our framework. On the impact side, Grassucci et al.~\cite{grassucci2026closing} show the gap disrupts semantic consistency, Ramasinghe et al.~\cite{ramasinghe2024accept} argue a moderate gap can benefit retrieval, and Yu et al.~\cite{yu2026modality} leverage it for MLLM scaling. These findings suggest the gap is task-dependent rather than uniformly harmful. However, none of these works propose an explicit decomposition into centroid and distributional components with quantitative evidence that the Distribution Gap is the true predictor of cross-modal task quality, nor do they offer a controllable mechanism to navigate this trade-off. Our work addresses this gap by providing both a principled decomposition and an effective controllable alignment framework.

\subsection{Mitigating the Modality Gap}

On the post-processing side, several methods attempt to close the gap without retraining. Mean-Centering~\cite{liang2022mind} subtracts modal centroids; GR-CLIP~\cite{grclip2025} and I0T~\cite{i0t2025} apply learned affine transforms; Yamashita et al.~\cite{yamashita2025bridging} calibrate the similarity distribution with pseudo-positive samples. These approaches are training-free but share a fundamental limitation: affine or translation-based operations can only move centroids closer, not reshape manifold geometry. On the training side, methods that modify model weights can in principle reshape feature distributions. Among our baselines, AlignCLIP~\cite{alignclip2025} trains from scratch with an alignment objective but struggles to recover pre-trained discriminative quality; M$^2$-Mix~\cite{oh2023geodesic} generates hard negatives via geodesic interpolation but does not directly target the gap; CLIP-Refine~\cite{cliprefine2025} proposes random-reference feature alignment whose stochastic directions lack a consistent alignment signal. Other recent efforts tackle the gap from diverse angles, including modality inversion~\cite{crossthegap2025}, diffusion-based bridging~\cite{diffusionbridge2025}, distributional divergence minimization~\cite{yin2025distributional}, softer contrastive objectives~\cite{fineclip2025}, and curriculum-based strategies~\cite{sofer2025pull}. Despite these advances, existing methods share two limitations: they do not distinguish centroid offset from distributional mismatch, and they produce a single fixed operating point rather than a controllable trade-off. TPC-CMA addresses both with a dual-mechanism loss, a gradient-aware curriculum schedule inspired by multi-task gradient balancing~\cite{chen2018gradnorm}, and a single user-controlled parameter $\alpha_{\text{target}}$ that governs the alignment strength.

\section{Conclusion \& Future Work}

By decomposing the modality gap into Centroid Gap and Distribution Gap, we show that Distribution Gap predicts cross-modal task quality far better than Raw Gap. Guided by this insight, TPC-CMA replaces cross-modal repulsion with intra-modal geometry matching and a three-phase curriculum for stable optimization. A single parameter $\alpha_{\text{target}}$ yields a controllable Pareto frontier: mild alignment preserves discriminative quality, while stronger alignment unlocks captioning and clustering. Spectral analysis reveals an intrinsic alignment-expressiveness tension: aggressive alignment triggers rank collapse, confirming that no single fixed operating point is universally optimal across tasks.

Future directions include automating $\alpha_{\text{target}}$ selection via validation-driven scheduling, designing rank-preserving constraints to push the Pareto frontier further, and extending to other modality pairs. Since the decomposition and curriculum are modality-agnostic, we expect TPC-CMA to readily transfer to audio-text and video-text settings with minimal architectural modification.

\bibliographystyle{plainnat}
\bibliography{sample-base}

\end{document}